\documentclass[letterpaper]{article} 
\usepackage{aaai2026}  
\usepackage{times}  
\usepackage{helvet}  
\usepackage{courier}  
\usepackage[hyphens]{url}  
\usepackage{graphicx} 
\usepackage{natbib}  
\usepackage{caption} 
\usepackage{algorithm}
\usepackage{multirow}
\usepackage{colortbl}
\usepackage{algorithm}
\usepackage{algorithmic}
\usepackage{caption}
\usepackage{booktabs, multirow, xcolor, colortbl}
\usepackage[table,xcdraw,dvipsnames]{xcolor}
\newcommand{\bftab}{\fontseries{b}\selectfont}
\usepackage{tikz}
\usepackage{adjustbox}
 \usepackage{amssymb}
 \usepackage{graphicx}
 \usepackage{amsmath}
 \usepackage{bm}
 \usepackage{amsthm}
 \usepackage{mathrsfs}
 \usepackage{array}
 \usepackage{booktabs}

\usepackage{newfloat}
\usepackage{listings}
\DeclareCaptionStyle{ruled}{labelfont=normalfont,labelsep=colon,strut=off} 
\floatstyle{ruled}
\newfloat{listing}{tb}{lst}{}
\floatname{listing}{Listing}
\title{MyGram: Modality-aware Graph Transformer with Global Distribution for \\Multi-modal Entity Alignment}

\author {
    Zhifei Li\textsuperscript{\rm 1,\rm 5,\rm 6},
    Ziyue Qin\textsuperscript{\rm 1,}\thanks{Corresponding Authors.},
    Xiangyu Luo\textsuperscript{\rm 4}, 
    Xiaoju Hou\textsuperscript{\rm 2}, \\
    Yue Zhao\textsuperscript{\rm 3,}\footnotemark[1], 
    Miao Zhang\textsuperscript{\rm 1}, 
    Zhifang Huang\textsuperscript{\rm 1},   
    Kui Xiao\textsuperscript{\rm 1}, 
    Bing Yang\textsuperscript{\rm 1,}\footnotemark[1]
}
\affiliations {
    \textsuperscript{\rm 1}School of Computer Science, Hubei University, Wuhan 430062, China\\
    \textsuperscript{\rm 2}Institute of Vocational Education, Guangdong Industry Polytechnic University, Guangzhou 510300, China\\
    \textsuperscript{\rm 3}Shandong Police College, Ji’nan 250200, China\\
    \textsuperscript{\rm 4}School of Cyber Science and Technology, Hubei University, Wuhan 430062, China\\
    \textsuperscript{\rm 5}Hubei Key Laboratory of Big Data Intelligent Analysis and Application (Hubei University), Wuhan 430062, China\\
    \textsuperscript{\rm 6}Key Laboratory of Intelligent Sensing System and Security (Hubei University), Ministry of Education, Wuhan 430062, China\\
    \{zhifei1993, zhangmiao, 20160006, xiaokui\}@hubu.edu.cn, 2023030010@gdip.edu.cn,\\
    zhaoy@sdpc.edu.cn, \{qinziyue, xyluo\}@stu.hubu.edu.cn, yangbing@126.com

}

\usepackage{bibentry}

\begin{document}

\maketitle

\begin{abstract}
Multi-modal entity alignment aims to identify equivalent entities between two multi-modal Knowledge graphs by integrating multi-modal data, such as images and text, to enrich the semantic representations of entities. However, existing methods may overlook the structural contextual information within each modality, making them vulnerable to interference from shallow features. To address these challenges, we propose MyGram, a \textbf{m}odalit\textbf{y}-aware \textbf{gra}ph transformer with global distribution for \textbf{m}ulti-modal entity alignment. Specifically, we develop a modality diffusion learning module to capture deep structural contextual information within modalities and enable fine-grained multi-modal fusion. In addition, we introduce a Gram Loss that acts as a regularization constraint by minimizing the volume of a 4-dimensional parallelotope formed by multi-modal features, thereby achieving global distribution consistency across modalities. We conduct experiments on five public datasets. Results show that MyGram outperforms baseline models, achieving a maximum improvement of 4.8\% in Hits@1 on FBDB15K, 9.9\% on FBYG15K, and 4.3\% on DBP15K.
\end{abstract}

\begin{links}
    \link{Code}{https://github.com/HubuKG/MyGram}
\end{links}

\section{Introduction}

\begin{figure}[!t]
    \centering
    \includegraphics[width=1.0\linewidth]{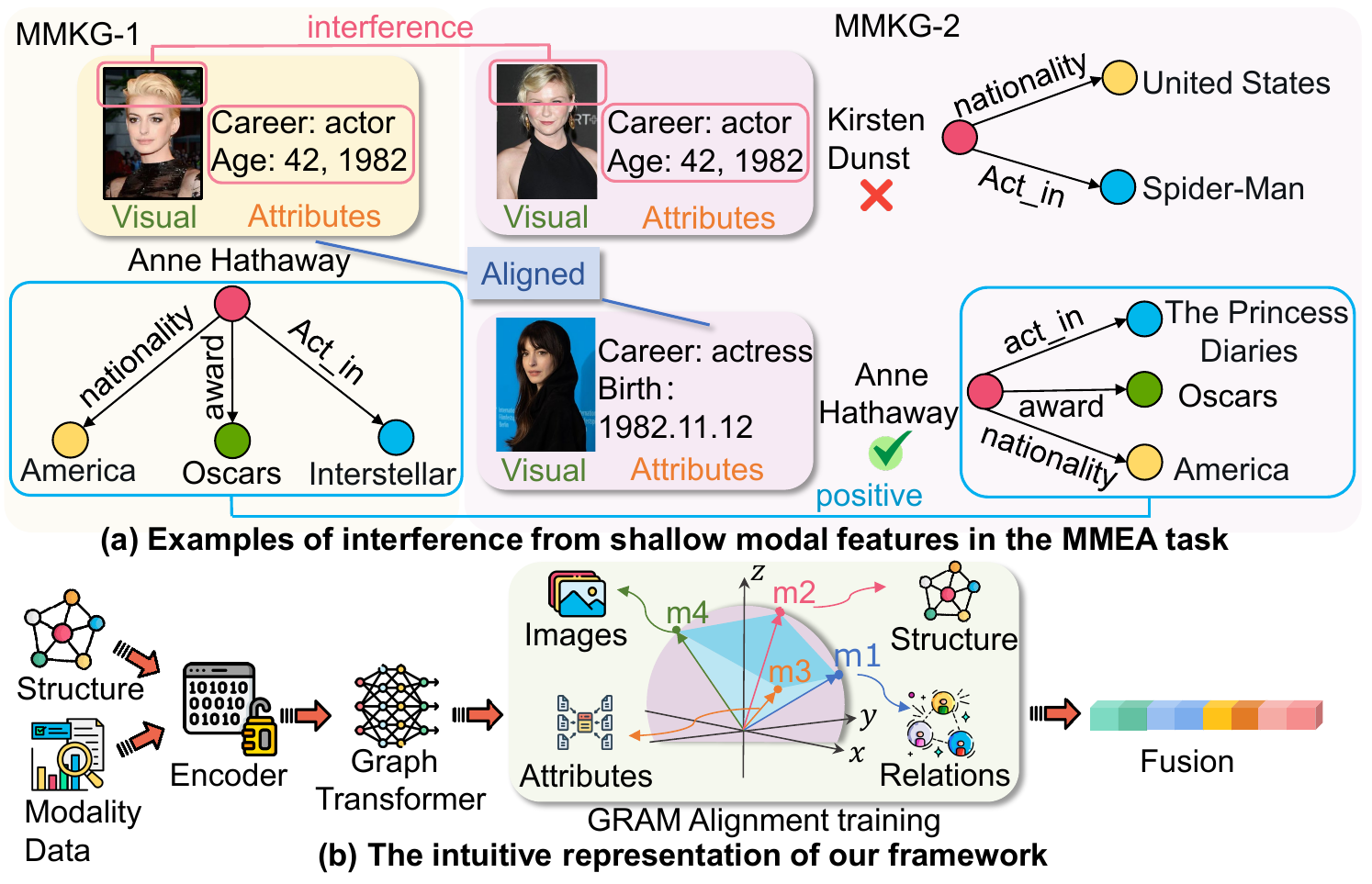}
    \caption{Illustration of modal interference and our GRAM-based alignment framework. (a) shows that when aligning the entity \texttt{Anne Hathaway}, the visual and attribute features of the entity \texttt{Kirsten Dunst} introduce interference to the task. (b) shows how Gram-based Loss imposes global constraints over cross-modal feature distributions.}
    \label{fig1}
\end{figure}

Knowledge graphs are structured graphical models for representing data \cite{1,39,40}. Since being introduced by Google in 2012, they have been widely adopted in applications such as intelligent question answering \cite{2,41,42}, recommendation systems \cite{3,45,46}, and various domains \cite{43,44}. A knowledge graph consists of entities and edges that denote the relationships between them, with each entity and relation associated with its attributes. At their core, they represent real-world knowledge in the form of (head entity, relation, tail entity) triples, enabling efficient data organization and management, as well as strong semantic representation and reasoning capabilities \cite{4,48}.

Multi-modal knowledge graphs (MMKGs) have attracted increasing attention from researchers \cite{5,47}. By integrating modalities such as text, images, and audio, MMKGs enhance the semantic representation of entities and improve the expressive power of knowledge graphs. However, due to heterogeneous data sources and differences in construction methods, MMKGs often present inconsistent representations of the same real-world entity. Thus, integrating MMKGs from multiple sources to ensure structural and semantic completeness has become a key research challenge \cite{7}. Among related tasks, multi-modal entity alignment (MMEA) plays a central role in knowledge fusion by identifying equivalent entities across MMKGs and establishing alignment links, thereby enabling the aggregation of cross-source knowledge \cite{6,8}.

Despite recent progress in MMEA, significant challenges remain in effectively modeling the cross-modal consistency of equivalent entities. (1) \textbf{Limitations of contrastive learning}. Most existing methods adopt intra-modal contrastive learning frameworks, aligning entities by optimizing the feature distances between positive and negative pairs. However, these approaches may overlook the distributional differences across modalities in the global feature space, which hinders cross-modal feature consistency. (2) \textbf{Interference from shallow features}. Existing methods overlook the structural contextual information within each modality, making it difficult for the model to distinguish between entities that appear similar but are essentially different. As illustrated in Figure \ref {fig1}(a), \texttt{Anne Hathaway} and \texttt{Kirsten Dunst} share highly similar visual and attribute features, which can introduce interference in the alignment process. However, by leveraging structural information, accurate alignment is still achievable. Therefore, it is necessary to utilize deeper and more discriminative modality-specific structural features.

To address the above issues, we propose \textbf{MyGram}, a modality-aware graph transformer with global distribution for multi-modal entity alignment. The model mainly includes the following two strategies: (1) \textbf{Gram-based Distribution Alignment.} As shown in Figure \ref{fig1}(b), we introduce Gram Loss as a regularization constraint, which minimizes the volume of a 4-dimensional parallelotope formed by modality vectors, thereby enhancing the global distribution consistency across modalities in the high-dimensional space. Compared to conventional point-wise feature alignment methods, Gram Loss promotes holistic cross-modal semantic coherence and improves the model's generalization ability. (2) \textbf{Modality Diffusion Learning}. To enrich entities' representation, we design a modality-aware graph convolutional diffusion module that captures multi-hop neighborhood information within each modality, thereby generating modality features enriched with structural context. On this basis, we further introduce a Transformer architecture that utilizes self-attention mechanisms to integrate information from various modalities, achieving deeper semantic fusion and more precise alignment. Overall, the main contributions of this paper are summarized as:
\begin{itemize}
    \item We propose a novel modality-aware framework, MyGram, which integrates graph diffusion and Transformer architectures to obtain structurally contextualized modality-specific features for more robust and reliable entity alignment.
    \item We introduce a Gram-based global alignment strategy that minimizes the volume of a 4-dimensional parallelotope formed by modality embeddings, enforcing global distribution alignment and improving the semantic consistency of equivalent entities.
    \item We perform extensive experiments on five separate datasets to validate the superiority of our model. For the Hits@1 metric, our model achieves a maximum improvement of 4.8\% on FBDB15K, 9.9\% on FBYG15K, and 4.3\% on DBP15K.
\end{itemize}

\section{Related Work}
This section provides a brief review of uni-modal entity alignment models and multi-modal entity alignment models.

\subsection{Uni-modal Entity Alignment}
Entity Alignment (EA) aims to detect semantically equivalent entities across distinct Knowledge Graphs (KGs), serving as a crucial step toward knowledge fusion \cite{9,10}. Traditional EA methods mainly target relational triple-based KGs and can be broadly grouped into: (1) Translation-based models, e.g., TransE \cite{1} and TransH \cite{2}, which embed entities and relations into vector spaces and capture relational semantics via translation. MTransE \cite{11} adds transition matrices for cross-graph mapping, while BootEA iteratively refines alignment through graph matching and embedding learning \cite{12}. (2) GNN-based models, which employ graph neural networks to model entity structures and enhance representations \cite{13,14}. GCN-Align \cite{15} applies graph convolutional networks, RDGCN \cite{16} models dual-graph structures, and MuGNN \cite{17} adopts multi-channel architectures for better alignment.

Uni-modal entity alignment approaches have shown effectiveness by leveraging embedding techniques to align entities. However, these methods may overlook valuable information from other modalities, such as textual descriptions and visual data, which can provide richer and more comprehensive representations of entities.

\subsection{Multi-modal Entity Alignment}
The emergence of MMKGs has drawn growing attention to incorporating visual and textual modalities into entity alignment \cite{18,19,20}. Recent studies focus on designing more effective multi-modal fusion mechanisms to improve alignment performance. MSNEA \cite{21} introduces visually guided relation and attribute learning to merge modality-specific features into a unified semantic representation. MoAlign \cite{22} employs hierarchical attention to capture structural, textual, and visual information. MEAformer \cite{23} adopts a dynamic cross-modal weighting strategy that adjusts modality contributions at the instance level in real time.

Furthermore, several MMEA approaches are proposed to address real-world constraints. To alleviate data scarcity, SimDiff applies diffusion-based augmentation for more stable learning \cite{24}. GSIEA \cite{25} reduces adverse structural discrepancies across KGs via graph structure prefix injection. PMF \cite{26} suppresses modality-irrelevant interference by progressively freezing each modality’s contribution during alignment. IBMEA \cite{27} introduces an information bottleneck to mitigate the influence of spurious cues.

Unlike prior methods, the proposed modality-aware graph convolutional diffusion Transformer captures high-order structural semantics and applies Gram-based distribution alignment to enhance cross-modal consistency. These strategies jointly improve robustness and alignment accuracy in multi-modal entity alignment tasks.

\begin{figure*}
    \centering
    \includegraphics[width=0.87\linewidth]{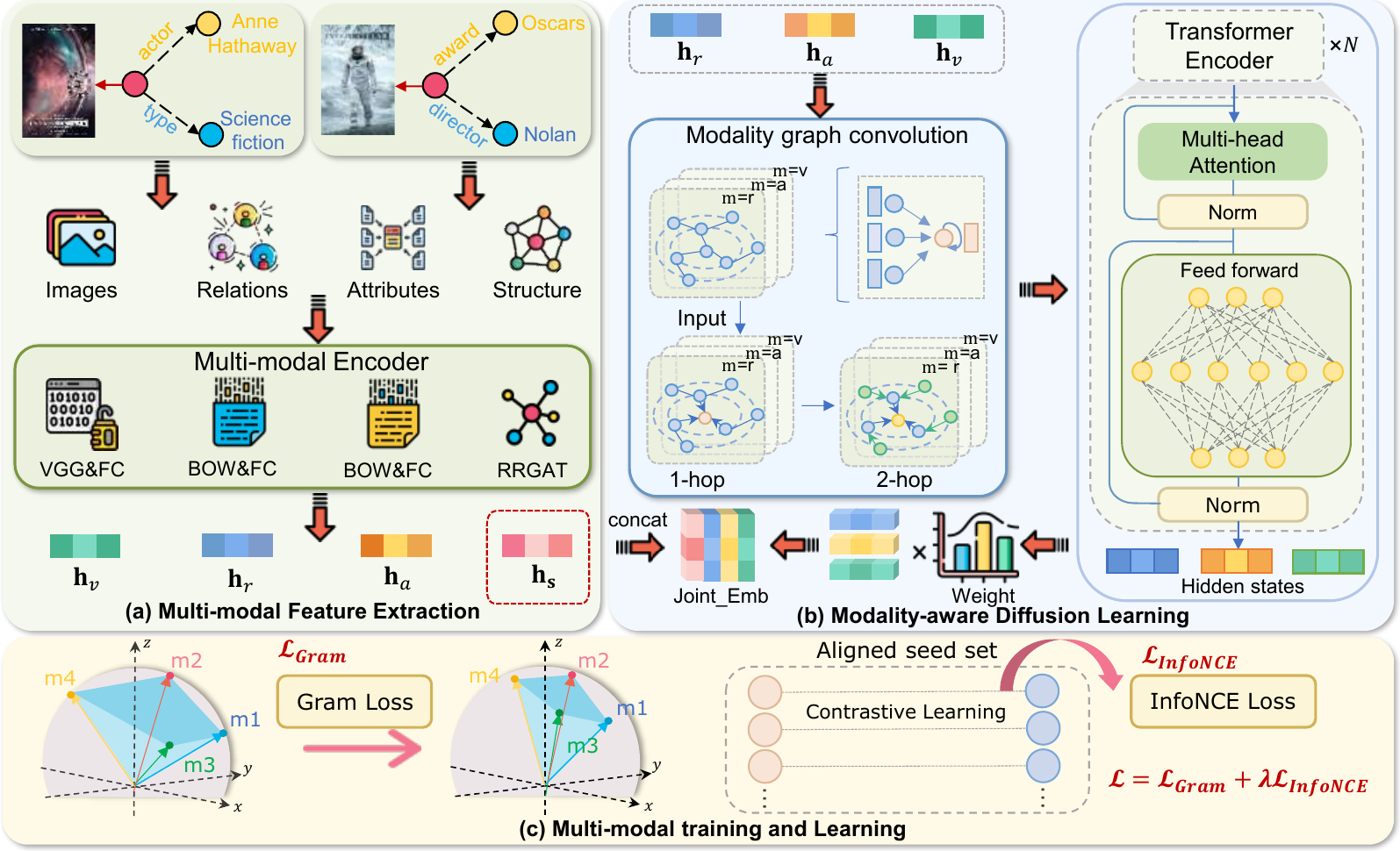}
    \caption{The overall framework of MyGram. (a) Multi-modal Feature Extraction: Extract uni-modal embeddings for each entity from different modalities; (b) Modality-aware Diffusion Learning: Enhance modality features with structural contextual information; (c) Multi-modal training and Learning: employing gram loss to establish alignment between equivalent entities.}
    \label{fig2}
\end{figure*}

\section{Methodology}

In this section, we present our MyGram framework, illustrated in Figure \ref{fig2}. Our model consists of three main modules: (1) Multi-modal Feature Extraction: extracting uni-modal embeddings from each entity; (2) Modality-aware Diffusion Learning: obtaining modality features with structural contextual information; (3) Multi-modal training and Learning: employing gram loss to establish alignment between equivalent entities.

\subsection{Preliminaries}
Multi-modal entity alignment seeks to identify entities from different knowledge graphs that refer to the same real-world object. Specifically, we consider two multi-modal knowledge graphs, denoted as 
$\mathcal{G}_1=\{\mathcal{E}_1,\mathcal{R}_1,\mathcal{A}_1,\mathcal{V}_1,\mathcal{T}_1\}$ and 
$\mathcal{G}_2=\{\mathcal{E}_2,\mathcal{R}_2,\mathcal{A}_2,\mathcal{V}_2,\mathcal{T}_2\}$, with $\mathcal{E},\mathcal{R},\mathcal{A}$, and $\mathcal{V}$ representing the set of entities, relations, attributes, and images, respectively. $\mathcal{T}$ signifies the set of triplets. MMEA aims to identify equivalent entity pairs $\mathcal{S}=\{(e_{1},e_{2})|e_{1}\in\mathcal{E}_{1},e_{2}\in\mathcal{E}_{2},$ $e_{1} \equiv e_{2}\}$, During training, the model is given a set of pre-aligned entity pairs $\mathcal{S}$. In the evaluation stage, for each given entity $e_{1}\in\mathcal{E}_{1}$, the model is expected to retrieve its corresponding equivalent entity $e_{2}\in\mathcal{E}_{2}$ from the full candidate set.

\subsection{Multi-modal Feature Extraction}
To extract features from different modalities, we extract entity features from MMKGs by processing each modality independently, thereby preserving its unique semantic information.

\textbf{Structure}
To capture the relationships between entities and their neighbors as well as the varying importance of the neighbors, we employ a relational reflection graph attention network (RRGAT) to aggregate the entity neighbors that retain the structural information of the relationships \cite{28}. Let ${x}_{g}\in \mathbb{R}^{d}$ represents the initialized entity feature, the feature of neighborhood aggregation is:
\begin{align}
    \mathbf{h}_{g} = RRGAT\left( {\omega,\mathbf{M}_{g},{x}_{g}} \right),
\end{align}
where $\omega$ is a learnable vector, $\mathbf{M}_{s}$ represents a relational transformation matrix that reflects the relationship.

\textbf{Relation, Attribute and Visual}
We first project the input features of relation $r$, attribute $a$, and visual $v$ modalities into a shared feature space, and obtain the modality embedding representation through linear transformation:
\begin{align}
    \mathbf{h}_{m} = ~\mathbf{W}_{m}x_{m} + ~b_{m},m \in \left\{ {r,a,v} \right\},
\end{align}
where $x_{m}$ is the initial feature of entity for modality $m$. $\mathbf{W}_{m}\in \mathbb{R}^{d \times d_m} $ and $\mathbf{b}_{m}$ denote the weight matrix and learnable parameter for the modality $m$, respectively. For attributes and relations, we represent them using bag-of-words features. For images, we preprocess its visual image using a pretrained image encoder and take the output from the encoder’s final layer as the image feature:
\begin{align}
    x_{v} = {ImageEncoder}\left( v \right).
\end{align}

\subsection{Modality-aware Diffusion Learning}
To address the challenge of interference from shallow features and ensure effective multi-modal fusion, we propose a novel \textbf{m}odality-aware \textbf{g}raph convolutional \textbf{d}iffusion module (MGD). While traditional methods may neglect the modality information of neighboring entities, our approach captures high-order neighbor features across modalities to obtain modality-specific representations enriched with structural context. This is followed by a Transformer-based self-attention mechanism that enables deeper semantic alignment and more comprehensive multi-modal fusion \cite{29}.

For the relation, attribute, and visual embeddings obtained in the previous section, we apply MGD to each modality individually. For each modality, we first construct a shared adjacency matrix with self-loops:
\begin{align}
    \hat{A} = ~D^{- \frac{1}{2}}\left( {A + I} \right)D^{- \frac{1}{2}},
\end{align}
where $A$ is the original adjacency matrix, $I$ is the identity matrix (used to introduce self-loops), and $D$ is the degree matrix computed by $D_{ii} = \sum_j (A + I)_{ij}$, where $D_{ii}$ represents the degree of node $i$.

After constructing the adjacency matrix $\hat{A}$, we process the input features of each modality as follows:
\begin{align}
    \mathbf{H}_{m}^{(l)} = ~\beta \cdot \hat{A}\mathbf{H}_{m}^{({l - 1})} + ~\alpha \cdot \mathbf{H}_{m}^{(0)}~~(l = 1,2,\cdots,k),
\end{align}
where $\mathbf{H}_{m,i}^{(0)}$ is the result of applying dropout to $\mathbf{h}_{i}^{m}$, $\beta$ denotes the neighborhood propagation coefficient, $\alpha$ represents the residual retention coefficient, and $H_{m,i}^{(l)}$ is the result of the $l$-th propagation. The final output is obtained after $k$-th propagation as follows, where $\gamma$ is a stabilization factor utilized for preventing gradient explosion:
\begin{align}
    \mathbf{H}_{m} = Dropout\left( \frac{1}{\gamma}\mathbf{H}_{m}^{(k)} \right), \gamma = ~\beta^{k} + ~\alpha{\sum\limits_{c = 0}^{k - 1}\beta^{c}}.
\end{align}

After obtaining modality features enriched with structural context, we further introduce the transformer self-attention mechanism to achieve multi-modal interactive fusion. Specifically, the transformer can effectively capture long-range dependencies across different modalities. Through self-attention, it dynamically adjusts the weights of modality features, thereby facilitating comprehensive interaction and fusion among modalities.

First, we apply cross-modal attention to each set of modality features $\mathbf{H}_m$ to obtain the attention weights:
\begin{align}
head_m^{\; i} = \beta_m^{(i)} V_m^{(i)}, \quad \beta_m = \mathrm{softmax}\left( \frac{Q_m^\top K_m}{\sqrt{d_h}} \right),
\end{align}
where $\mathbf{Q}_m$, $\mathbf{K}_m$, and $\mathbf{V}_m^{(i)}$ denote the query, key, and value matrices for the $i$-th attention head, and $d_h$ is the dimension of each head. Then we employ multi-head cross-attention (\textit{MA}) to enhance the ability to capture diverse information across modalities:
\begin{align}
    MA\left( \mathbf{H}_{m} \right) = \left\lbrack {{head}_{m}^{\;1} \oplus ,\cdots, \oplus {head}_{m}^{N_{h}}} \right\rbrack\mathbf{W}_{o} ~,
\end{align}
where $\oplus$ denotes the concatenation operation, $N_h$ is the number of attention heads, and $\mathbf{W}_o$ is the output projection matrix.

After processing, residual links, normalization, and feed forward are applied, ultimately resulting in the hidden state ${\widetilde{\mathbf{H}}}_{m}$. Finally, we define the cross-modal weights for each modality to achieve multi-modal fusion:
\begin{align}
    \omega_{m} = ~\frac{exp\left( ~{\sum\limits_{j \in M}{\sum\limits_{i = 0}^{N_{h}}\beta_{m,j}^{(i)}}}/\sqrt{|M| \times N_{h}} \right)}{\sum\limits_{k \in M}{exp\left( ~{\sum\limits_{k \in M}{\sum\limits_{i = 0}^{N_{h}}\beta_{m,k}^{(i)}}}/\sqrt{|M| \times N_{h}} \right)}}.
\end{align}

Next, we define the joint embedding for modality fusion:
\begin{align}
    \mathbf{H}_{o} = ~{\mathbf{H}_{g} \oplus}_{m \in M}\left\lbrack {\omega_{m}\mathbf{H}_{m}} \right\rbrack.
\end{align} 

\subsection{Multi-modal training and Learning}
Most existing methods adopt contrastive learning during training. They impose distance constraints between the features of positive and negative entity pairs to promote accurate entity alignment.
However, such methods may ignore the geometric relationships among vectors in high-dimensional space, making it difficult to fully capture deeper semantic consistency across modalities.

To address the above issue, we draw inspiration from \cite{30}, who propose using the volume formed by multiple modality vectors in high-dimensional space as a geometric indicator of vector relationships. This volume offers a more intuitive way to reflect the consistency of multi-modal features in high-dimensional space. We apply this design to multi-modal entity alignment, where introducing high-dimensional volume as a regularization constraint helps enforce semantic consistency between different modality features of the same entity from a geometric perspective. A smaller volume implies that the embeddings lie in a more compact subspace, thereby indicating stronger semantic coherence across modalities.

During the training process, we first compute the similarity matrix between the structural feature of the source entity and the visual feature of the target entity in each entity pair. The multi-modal features are represented by the hidden states ${\widetilde{\mathbf{H}}}_{m}$ obtained in the previous section:
\begin{align}
\text{sim} = \frac{\langle \widetilde{\mathbf{H}}_{g}^{s}, \widetilde{\mathbf{H}}_{v}^{t} \rangle}{\|\widetilde{\mathbf{H}}_{g}^{t}\| \cdot \|\widetilde{\mathbf{H}}_{v}^{t}\|}~.
\end{align}

Based on the computed similarity matrix, we select the top-K most similar candidate entities for each source entity.
Then, we built a 4-dimensional parallelotope using the structural feature of the source entity and the visual, attribute, and relation features of the target entity in each entity pair. 
\begin{align}
    \mathcal{M} = \left\lbrack {{\widetilde{\mathbf{H}}}_{g}^{s},{\widetilde{\mathbf{H}}}_{v}^{t},{\widetilde{\mathbf{H}}}_{a}^{t},{\widetilde{\mathbf{H}}}_{r}^{t}} \right\rbrack \in \mathbb{R}^{d_{h} \times 4},
\end{align}
where $\mathcal{M}$ represents the multi-modal matrix we construct. Then, the Gram matrix $G \in \mathbb{R}^{4 \times 4}$ is defined:
\begin{align}
G~ = \mathcal{M}^\top \mathcal{M} =
\begin{bmatrix}
\langle \widetilde{\mathbf{H}}_g^{s}, \widetilde{\mathbf{H}}_g^{s} \rangle & \cdots & \langle \widetilde{\mathbf{H}}_g^{t}, \widetilde{\mathbf{H}}_r^{t} \rangle \\
\vdots & \ddots & \vdots \\
\langle \widetilde{\mathbf{H}}_r^{t}, \widetilde{\mathbf{H}}_g^{s} \rangle & \cdots & \langle \widetilde{\mathbf{H}}_r^{t}, \widetilde{\mathbf{H}}_r^{t} \rangle
\end{bmatrix}.
\end{align}

According to \cite{31}, if $\widetilde{\mathbf{H}}_{g}, \ldots, \widetilde{\mathbf{H}}_{r}$ are vectors in $\mathbb{R}^{d_{h}}$ spanning a $4$-dimensional parallelotope, then the square of the volume of this shape is given by the determinant of the Gram matrix $G \in \mathbb{R}^{4 \times 4}$, known as the Gramian. From this, we obtain the volume of the 4-dimensional parallelotope as:
\begin{align}
    Vol = ~\sqrt{\left| {det\left( G~ \right)} \right| + ~\epsilon} ~.
\end{align}

To ensure that the correct positive match is included in the top-k candidate entities and to further locate its position within the top-k, we define a binary mask:
\begin{align}
    {mask}^{(i,k)} = ~\left\{ \begin{matrix}
{1~~~~if~{topk\_ idx}^{(i,k)} = {target}^{(i)}} \\
{~0~~~~~~~~~~~~~~~~~~~~~~~~~~~~~~~~~otherwise}
\end{matrix} \right.~~,
\end{align}
 where ${topk\_idx}^{(i,k)}$ denotes the global index of the $k$-th most similar neighbor of sample $i$. The mask is used to identify the correct position $p$ of the positive match, which is then used to extract its corresponding log-probability from the softmax distribution in the sparse contrastive loss:
\begin{align}
    \mathcal{L}_{Gram} = \mathcal{~} - \frac{1}{M}{\sum\limits_{m = 1}^{M}{log\frac{exp\left( - {Vol}^{(m,p)}/\tau \right)}{\sum\limits_{k = 1}^{K}{exp\left( - {Vol}^{(m,k)}/\tau \right)}}}},
\end{align}
where ${Vol}^{(i, k)}$ denotes the volume spanned by the structural embedding of the source entity and the multi-modal embedding of the target entity, and $\tau$ is the temperature coefficient. We further introduce a contrastive alignment loss, which aims to maximize the similarity of true aligned entity pairs and separate them from negative samples:
\begin{align}
    \mathcal{L}_{InfoNCE} = \mathcal{~}{\sum\limits_{(e_{i},e_{j}) \in \mathcal{S}}{- log\frac{exp\left( sim\left( e_{i},e_{j} \right)/\mathcal{T} \right)}{\sum\limits_{e_{k} \in \mathcal{N}_{i}^{neg}}{exp\left( sim\left( e_{i},e_{k} \right)/\mathcal{T} \right)}}}},
\end{align}
where $\text{sim}(e_i, e_j)$ represents the similarity (e.g., cosine similarity) between the source entity $e_i$ and the target entity $e_j$, and $\mathcal{T}$ is the temperature coefficient. For each aligned entity pair $(e_i, e_j)$, we treat it as a positive sample, while sampling multiple negative entities from the candidate set $\mathcal{N}_i$ to construct a local contrastive learning objective. The final loss is formulated as a weighted combination of the two components, $\lambda$ is used to adjust the weight of gram loss:
\begin{align}
    \mathcal{L}_{total} = \mathcal{L}_{InfoNCE} + \lambda \mathcal{L}_{Gram}.
\end{align}

\begin{table*}[!t]
\centering
\setlength{\tabcolsep}{2.5pt}
\renewcommand{\arraystretch}{1.}
\resizebox{\textwidth}{!}{%
\begin{tabular}{l|ccc|ccc|ccc|ccc|ccc|ccc}
\toprule[1.2pt]
\multirow{3}{*}{\bftab Model} 
  & \multicolumn{9}{c|}{\cellcolor{CornflowerBlue!5}\textbf{FB15K-DB15K}} 
  & \multicolumn{9}{c}{\cellcolor{LimeGreen!5}\textbf{FB15K-YG15K}} \\ 
\cmidrule(lr){2-10} \cmidrule(lr){11-19}
 & \multicolumn{3}{c|}{$R_{Seed}=20\%$} & \multicolumn{3}{c|}{$R_{Seed}=50\%$} & \multicolumn{3}{c|}{$R_{Seed}=80\%$}
 & \multicolumn{3}{c|}{$R_{Seed}=20\%$} & \multicolumn{3}{c|}{$R_{Seed}=50\%$} & \multicolumn{3}{c}{$R_{Seed}=80\%$} \\ 
\cmidrule(lr){2-4} \cmidrule(lr){5-7} \cmidrule(lr){8-10}
\cmidrule(lr){11-13} \cmidrule(lr){14-16} \cmidrule(lr){17-19}
& \scriptsize{MRR} & \scriptsize{Hit@1} & \scriptsize{Hit@10}
& \scriptsize{MRR} & \scriptsize{Hit@1} & \scriptsize{Hit@10}
& \scriptsize{MRR} & \scriptsize{Hit@1} & \scriptsize{Hit@10}
& \scriptsize{MRR} & \scriptsize{Hit@1} & \scriptsize{Hit@10}
& \scriptsize{MRR} & \scriptsize{Hit@1} & \scriptsize{Hit@10}
& \scriptsize{MRR} & \scriptsize{Hit@1} & \scriptsize{Hit@10} \\ 
\midrule
PoE         & .170 & .126 & .251 & .533 & .464 & .658 & .721 & .666 & .820 & .154 & .113 & .229 & .414 & .347 & .536 & .635 & .573 & .746 \\
MMEA        & .357 & .265 & .541 & .512 & .416 & .703 & .685 & .590 & .868 & .317 & .234 & .480 & .486 & .403 & .645 & .682 & .597 & .839 \\
MSNEA       & .175 & .114 & .296 & .388 & .288 & .590 & .613 & .518 & .779 & .153 & .103 & .249 & .413 & .320 & .589 & .620 & .531 & .778 \\
MCLEA       & .393 & .295 & .582 & .652 & .573 & .800 & .784 & .730 & .883 & .332 & .254 & .484 & .616 & .543 & .759 & .715 & .653 & .835 \\
ACK-MMEA    & .387 & .304 & .549 & .624 & .560 & .736 & .752 & .682 & .874 & .360 & .289 & .496 & .593 & .535 & .699 & .744 & .676 & .864 \\
MoAlign     & .409 & .318 & .564 & .634 & .576 & .749 & .773 & .699 & .882 & .378 & .296 & .525 & .617 & .550 & .713 & .769 & .689 & .884 \\
MEAformer   & .534 & .434 & .728 & .704 & .625 & .847 & .825 & .773 & .918 & .416 & .325 & .598 & .640 & .560 & .780 & .768 & .705 & .874 \\
GEEA        & .450 & .343 & .661 & .723 & .651 & .852 & .836 & .787 & .918 & .393 & .298 & .585 & .668 & .589 & .794 & .780 & .732 & .890 \\
SimDiff     & .678 & .615 & \underline{\cellcolor{gray!15}.820}& .786 & .731 & \underline{\cellcolor{gray!15}.880} & \underline{\cellcolor{gray!15}.865} & \underline{\cellcolor{gray!15}.829} & \underline{\cellcolor{gray!15}.929} & \underline{\cellcolor{gray!15}.595} & \underline{\cellcolor{gray!15}.530} & \underline{\cellcolor{gray!15}.736} & \underline{\cellcolor{gray!15}.716} & \underline{\cellcolor{gray!15}.659} & .820 & .791 & .743 & .886 \\
DESAlign    & .586 & .497 & .750 & .728 & .656 & .853 & .850 & .805 & .926 & .495 & .410 & .660 & .642 & .573 & .763 & .782 & .728 & .877 \\
PMF         & .627 & .547 & .776 & .786 & .724 & .878 & .861 & .823 & .923 & .546 & .466 & .701 & .706 & .644 & .815 & \underline{\cellcolor{gray!15}.806} & \underline{\cellcolor{gray!15}.756} & .892 \\
IBMEA       & \underline{\cellcolor{gray!15}.697} & \underline{\cellcolor{gray!15}.631} & .813 & \underline{\cellcolor{gray!15}.793} & \underline{\cellcolor{gray!15}.742} & \underline{\cellcolor{gray!15}.880} & .859 & .821 & .922 & .584 & .521 & .708 & .714 & .655 & \underline{\cellcolor{gray!15}.821} & .800 & .751 & .890 \\
SNAG        & .495 & .389 & .702 & .742 & .669 & .875 & .848 & .802 & .928 & .405 & .309 & .596 & .658 & .578 & .804 & .804 & .747 & \underline{\cellcolor{gray!15}.902} \\
GSIEA       & .547 & .458 & .715 & .736 & .669 & .855 & .844 & .801 & .923 & .468 & .380 & .637 & .676 & .604 & .806 & .789 & .735 & .892 \\ 
\midrule
MyGram & \cellcolor{pink!20}\bftab.739 & \cellcolor{pink!20}\bftab.679 
& \cellcolor{pink!20}\bftab.849 & \cellcolor{pink!20}\bftab.813 
& \cellcolor{pink!20}\bftab.764 & \cellcolor{pink!20}\bftab.907 
& \cellcolor{pink!20}\bftab.879 & \cellcolor{pink!20}\bftab.842 
& \cellcolor{pink!20}\bftab.948 & \cellcolor{pink!20}\bftab.693 
& \cellcolor{pink!20}\bftab.629 & \cellcolor{pink!20}\bftab.820 
& \cellcolor{pink!20}\bftab.771 & \cellcolor{pink!20}\bftab.715 
& \cellcolor{pink!20}\bftab.884 & \cellcolor{pink!20}\bftab.836 
& \cellcolor{pink!20}\bftab.783 & \cellcolor{pink!20}\bftab.938 \\ 
\bottomrule[1.2pt]
\end{tabular}%
}
\caption{Performance comparison of different MMEA models on FBDB15K and FBYG15K. The optimal results are highlighted in \textbf{bold}, while the second-best results are \underline{underlined}.}
\label{tab1}
\end{table*}
\section{Experiments}
To evaluate the effectiveness of the proposed MyGram model, we conduct a thorough experimental study on six subsets from two benchmark datasets. This evaluation is designed to investigate the following five research questions.

\textbf{RQ1:} How does the performance of MyGram compare to other MMEA models?

\textbf{RQ2:} How does modal information impact the performance of MyGram? 

\textbf{RQ3:} How do the principal modules affect the performance of MyGram?

\textbf{RQ4:} How does MyGram perform with a low alignment pair rate?

\textbf{RQ5:} How does MyGram perform when applied to real-world MMEA tasks?

\subsection{Experimental Settings}

\subsubsection{Datasets Statistics:}
To assess the effectiveness of our proposed method, we adopt two widely used types of MMEA datasets: cross-knowledge graph datasets and bilingual datasets. For the cross-KG setting, we select FB15K-DB15K and FB15K-YAGO15K \cite{5}. For the bilingual setting, we employ the DBP15K dataset \cite{32}. Following previous studies, we utilize 20\%, 50\%, and 80\% of the alignment pairs as training seeds for cross-KG datasets, and 30\% for bilingual datasets. For entities lacking associated images, we assign randomly initialized vectors for the visual modality, consistent with the settings adopted in prior works \cite{33}.

\subsubsection{Evaluation Metrics:}
Evaluation is conducted using standard metrics, including Hits@N (where N = 1, 10) and Mean Reciprocal Rank (MRR). Hits@N indicates the proportion of correct entities ranked in the top N, while MRR (Mean Reciprocal Rank) represents the average reciprocal rank of the correct entities. Higher values of Hits@N and
MRR indicate better performance.

\begin{table}[!h]
\centering
\renewcommand{\arraystretch}{1.0}
\setlength{\tabcolsep}{6pt}
\resizebox{0.92\columnwidth}{!}{%
\begin{tabular}{c|c|c|c|c}
\toprule[1.2pt]
\textbf{Datasets} & \textbf{Models} & \textbf{MRR} & \textbf{Hit@1} & \textbf{Hit@10} \\ 
\midrule
\multirow{6}{*}{\rotatebox{90}{DBP15K$_{\text{ZH-EN}}$}} 
    & MSNEA & .684 & .601 & .830 \\
    & MCLEA & .788 & .715 & .923 \\
    & MEAformer & .835 & .771 & .951 \\
    & DESAlign & \underline{\cellcolor{gray!15}.865} & \underline{\cellcolor{gray!15}.810} & \underline{\cellcolor{gray!15}.957} \\
    & GSIEA & .855 & .786 & .952 \\  
    & \textbf{MyGram} & \cellcolor{pink!20}\bftab .876 & \cellcolor{pink!20}\bftab .833 & \cellcolor{pink!20}\bftab .960 \\
\midrule
\multirow{6}{*}{\rotatebox{90}{DBP15K\textsubscript{JA-EN}}} 
    & MSNEA & .617 & .535 & .775 \\
    & MCLEA & .785 & .715 & .909 \\
    & MEAformer & .834 & .764 & .959 \\
    & DESAlign & \underline{\cellcolor{gray!15}.869} & \underline{\cellcolor{gray!15}.811} & \underline{\cellcolor{gray!15}.963} \\
    & GSIEA & .852 & .787 & .962 \\ 
    & \textbf{MyGram} & \cellcolor{pink!20}\bftab .879 & \cellcolor{pink!20}\bftab .836 & \cellcolor{pink!20}\bftab .964 \\
\midrule
\multirow{6}{*}{\rotatebox{90}{DBP15K\textsubscript{FR-EN}}} 
    & MSNEA & .630 & .543 & .801 \\
    & MCLEA & .782 & .711 & .909 \\
    & MEAformer & .841 & .772 & .962 \\
    & DESAlign & \underline{\cellcolor{gray!15}.885} & \underline{\cellcolor{gray!15}.826} & \underline{\cellcolor{gray!15}.972} \\
    & GSIEA & .865 & .796 & .968 \\  
    & \textbf{MyGram} & \cellcolor{pink!20}\bftab .908 & \cellcolor{pink!20}\bftab .869 & \cellcolor{pink!20}\bftab .979 \\
\bottomrule[1.2pt]
\end{tabular}%
}
\caption{Performance comparison of different MMEA models on DBP15K.}
\label{tab2}
\end{table}
\subsubsection{Baseline Models:}
To verify the effectiveness of the proposed method, we perform a comprehensive comparison against a set of representative and competitive MMEA models: PoE \cite{5}, MMEA \cite{8}, MSNEA \cite{21}, MCLEA \cite{34}, ACK-MMEA \cite{35}, MoAlign \cite{22}, MEAformer \cite{23}, GEEA \cite{36}, SimDiff \cite{24}, DESAlign \cite{37}, PMF \cite{26}, IBMEA \cite{27}, SNAG \cite{38}, GSIEA \cite{25}.

\subsubsection{Parameter Settings:}
We standardize the hidden layer size across all network components to 300 dimensions. The training process is conducted for a total of 1000 epochs, with the learning rate initialized at 5e-3. For the multi-modal embedding module, we adopt VGG-16 as the image feature extractor, setting the visual embedding $\mathbf{d}_{v}$ to 4096. Regarding the transformer component within our model, the intermediate layer dimension is configured to 400, and the number of self-attention heads is set to 5.

\subsection{Performance Comparison (RQ1)}
Tables ~\ref{tab1} and ~\ref{tab2} show the experimental results of the multi-modal entity alignment task on two monolingual datasets FBDB15K, FBYG15K, as well as the bilingual dataset DBP15K, under an iterative setting. As anticipated, the proposed MyGram model demonstrates superior performance in all metrics across the nine benchmark experiments when compared to existing state-of-the-art models. Specifically, compared to the second-best performing models, MyGram exhibits average improvements of 4.8\%, 9.9\%, and 4.3\% in the Hit@1 metric on the FBDB15K, FBYG15K, and DBP15K datasets, respectively. 

The experimental results demonstrate that our proposed MyGram model not only significantly improves the accuracy of multi-modal entity alignment but also enhances its generalization capability across a variety of datasets. Notably, MyGram outperforms the state-of-the-art methods SimDiff and IBMEA. MyGram proposes a modality graph convolutional diffusion method to introduce contextual semantic information for multi-modal features. In addition, it utilizes a specially designed Gram alignment constraint loss that promotes inter-modal semantic structure consistency to improve model robustness. The superior performance of MyGram compared to other multi-modal models suggests promising research directions for utilizing the neighborhood context of multi-modal features in the MMEA task.

\begin{table}[]
\centering
\setlength{\tabcolsep}{1.5pt} 
\renewcommand{\arraystretch}{1.0} 
\resizebox{\columnwidth}{!}{%
\begin{tabular}{l|ccc| ccc}
\toprule[1.1pt]
\multirow{2}{*}{Model} & \multicolumn{3}{c|}{FB15K-DB15K} & \multicolumn{3}{c}{FB15K-YG15K} \\
\cmidrule(lr){2-4} \cmidrule(lr){5-7}
 & MRR & Hit@1 & Hit@10 & MRR & Hit@1 & Hit@10 \\
\midrule
w/o relation    & .842 & .822 & .934 & .811 & .761 & .922 \\
w/o attributes  & \underline{\cellcolor{gray!15}.859} & \underline{\cellcolor{gray!15}.834} & .940 & .818 & .768 & .925 \\
w/o image       & .851 & .829 & \underline{\cellcolor{gray!15}.942} & \underline{\cellcolor{gray!15}.824} & \underline{\cellcolor{gray!15}.772} & \underline{\cellcolor{gray!15}.927} \\
\midrule
MyGram            & \cellcolor{pink!20}\textbf{.879} & \cellcolor{pink!20}\textbf{.842} & \cellcolor{pink!20}\textbf{.948} & \cellcolor{pink!20}\textbf{.836} & \cellcolor{pink!20}\textbf{.783} & \cellcolor{pink!20}\textbf{.938} \\
\bottomrule[1.1pt]
\end{tabular}
}
\caption{Ablation study for modalities with 80\% seed.}
\label{tab3}
\end{table}

\subsection{Modality Effect (RQ2)}
In order to comprehensively analyze the effects of various factors on the model performance, we design ablation experiments for the modalities. Specifically, we consider three variants with missing modalities: w/o Relation, w/o Attribute, and w/o Image, which denote the model with relation, attribute, and visual modality of the entity removed, respectively. The ablation results on FBDB15K and FBYG15K datasets are presented in Table~\ref {tab3}.

Experimental results on both datasets show that the model achieves the best performance when all modal features are used for entity representation. Removing any single modality leads to performance degradation to varying degrees. In particular, excluding the relational modality results in the most significant performance drop across both datasets, indicating the important role of relational information in multimodal entity alignment. Moreover, the results of other variants further validate the effectiveness of the model in leveraging multi-modal information for entity alignment.

\subsection{Key Components (RQ3)}
\begin{figure}[t]
    \centering
    \includegraphics[width=0.98\linewidth]{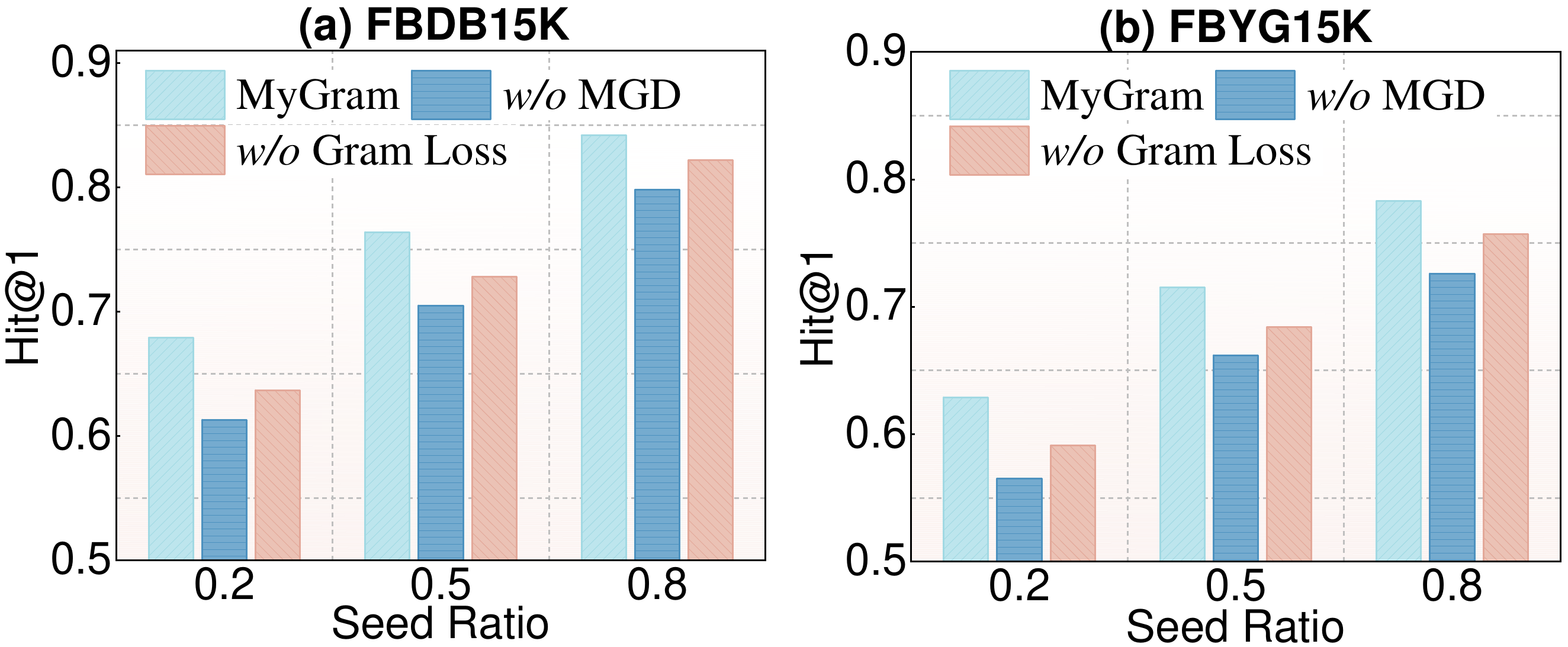}
    \caption{Ablation study on key components of proposed MyGram on (a) FBDB15K and (b) FBYG15K.}
    \label{fig3}
\end{figure}

Analogously, we design two variants for the key components ablation study of the proposed model: (1) w/o MGD: a variant that removes the modality-aware graph convoluational diffusion module; (2) w/o Gram: a variant with Gram-based loss removed. Figure~\ref{fig3} presents the impact of these variations on the model.

For the key components of the model, the model performance is significantly degraded in the absence of the modal map convolutional diffusion module. In addition, removing the Gram-based contrast loss also affects the performance. These results validate the effectiveness of the modal graph convolutional diffusion approach and the strategy for promoting multi-modal semantic coherence studied in this paper. These components enhance semantic propagation by taking into account the context of each modal neighborhood and strengthen the semantic consistency between the different modalities of an entity, demonstrating the effective role of modal semantics for MMEA tasks.

\begin{figure}[t]
    \centering
    \includegraphics[width=1.0\linewidth]{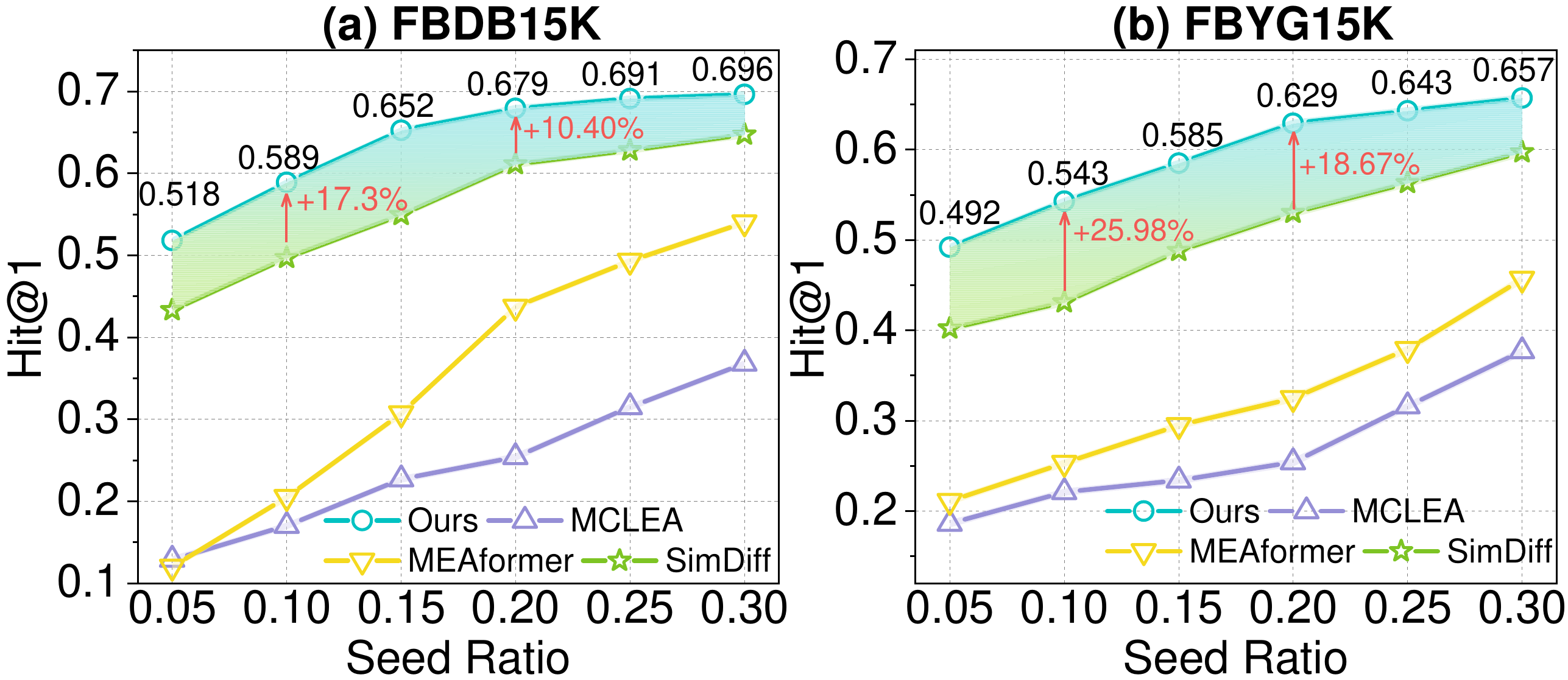}
    \caption{Low resource performance comparison on (a) FBDB15K and (b) FBYG15K.}
    \label{fig4}
\end{figure}

\subsection{Low-Resource Study (RQ4)}
To further evaluate the performance of MyGram in low-resource scenarios, we conduct experiments to assess its stability under conditions with an extremely low ratio of aligned seed pairs. In MMEA tasks, pre-aligned entity pairs are usually used to guide the training of the model, which can be viewed as a form of supervised learning. However, in real-world application scenarios, the availability of aligned seed pairs is rare or almost non-existent. Therefore, conducting low-resource training experiments can effectively analyze the robustness and generalization ability of the model in a weakly supervised or unsupervised environment. As illustrated in Figure ~\ref{fig4}, we compare the performance metrics of MyGram, MEAformer, and SimDiff across varying seed proportions, ranging from 5\% to 30\%. It can be clearly observed that as the alignment seed rate decreases, all the metrics of each model show performance degradation. However, our proposed MyGram still maintains the performance advantage over the other three models during the decrease in the resource ratio, demonstrating the significant effectiveness of our approach in low-resource scenarios.

\subsection{Case Study (RQ5)}
To evaluate the effectiveness of MyGram, we conducted a case study on the entity \texttt{Shang Hai} from the FB15K-DB15K dataset, as shown in Figure 4. In this case, the entities \texttt{Shang Hai} and \texttt{Hong Kong} share similar modality features and are easily affected by shallow feature interference. The structure of the MMEA task involves predicting the missing equivalent entity in a pair \texttt{(Shang Hai, ?)}. When provided with pre-aligned seed pairs as supervised examples to guide the training process, MyGram identifies the potential target entity of \texttt{Shang Hai} in DB15K and ranks the candidate entities accordingly. As shown in the prediction results, we observe that MEAformer and PMF assign a lower rank to the correct entity. In contrast, MyGram accurately identifies the correct match, demonstrating its superior ability to capture deeper information in the MMEA task.

\begin{figure}[]
    \centering
    \includegraphics[width=1.0\linewidth]{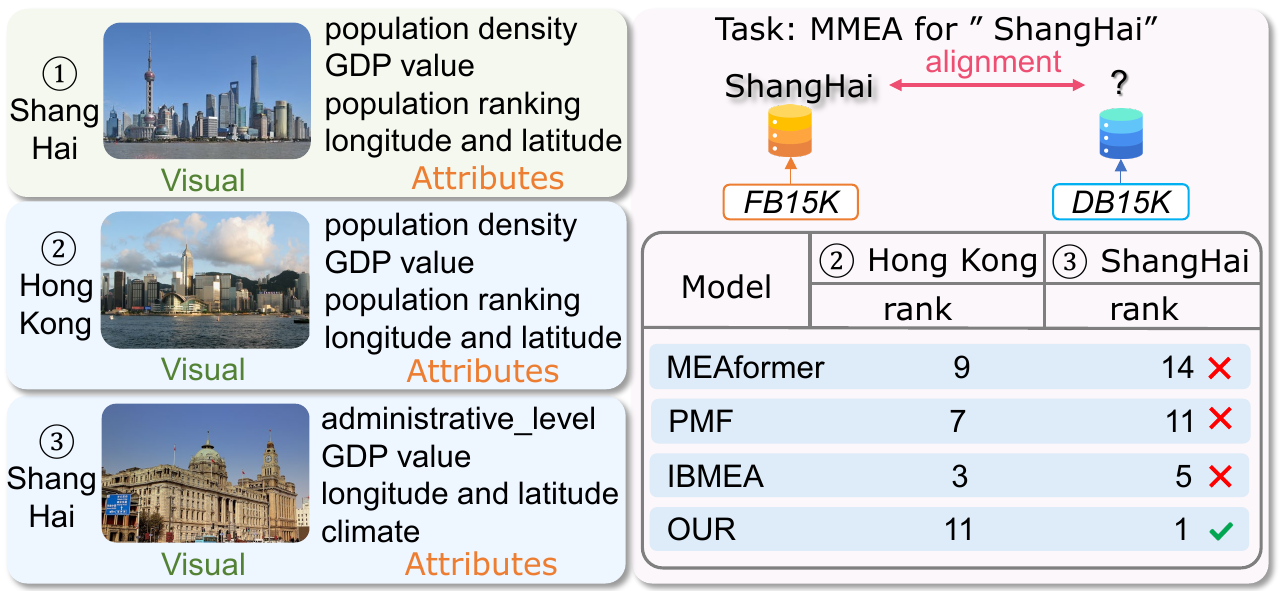}
    \caption{A case example of multi-modal entity alignment task for entity \texttt{Shang Hai}.}
    \label{fig5}
\end{figure}

\section{Conclusion}
In this paper, we propose MyGram for multi-modal entity alignment (MMEA). Our method studies the impact of shallow modality features on alignment and introduces Modality-aware Diffusion Learning to obtain modality representations enriched with structural context, effectively mitigating shallow feature interference. Moreover, MyGram incorporates a Gram-based loss to regularize cross-modal feature distributions, thus promoting global semantic consistency. Experiments show that MyGram consistently outperforms state-of-the-art methods on MMEA. In future work, we plan to explore integrating large language models (LLMs) to further enhance MMEA.

\section{Acknowledgments}
This work was supported in part by the National Natural Science Foundation of China (No. 62207011, 62407013, 62377009, 62101179), the Natural Science Foundation of Hubei Province of China (No. 2025AFB653), the Natural Science Foundation of Shandong Province of China (No. ZR2024QF257), the Science and Technology Support Plan for Youth Innovation of Colleges and Universities of Shandong Province of China (No. 2023KJ370), the Open Fund of Hubei Key Laboratory of Big Data Intelligent Analysis and Application, Hubei University (No. 2024BDIAA05), and the Open Fund of Key Laboratory of Intelligent Sensing System and Security of Hubei University, Ministry of Education (No. KLISSS202410).

\end{document}